\documentclass{article}
\usepackage{ijcai26}

\usepackage{times}
\usepackage{soul}
\usepackage{url}
\usepackage[hidelinks]{hyperref}
\usepackage[utf8]{inputenc}
\usepackage[small]{caption}
\usepackage{graphicx}
\usepackage{amsmath}
\usepackage{amsthm}
\usepackage{booktabs}
\usepackage{algorithm}
\usepackage{algorithmic}
\usepackage[switch]{lineno}

\usepackage{amssymb}
\usepackage{bm}
\usepackage{multirow}
\usepackage{array}
\usepackage[table]{xcolor}

\title{HomoFM: Deep Homography Estimation with Flow Matching}

\author{
Mengfan He$^1$\thanks{Equal contribution}\and
Liangzheng Sun$^2$\footnotemark[1]\and
Chunyu Li$^3$\And
Ziyang Meng$^1$\thanks{Corresponding author.}\\
\affiliations
$^1$Department of Precision Instrument, Tsinghua University, Beijing, China\\
$^2$School of Instrument Science and Opto-Electronics Engineering, Beijing Information Science and Technology University, Beijing, China\\
$^3$School of Aerospace Engineering, Beijing Institute of Technology, Beijing, China\\
\emails
}

\begin{document}

\maketitle

\begin{abstract}
Deep homography estimation has broad applications in computer vision and robotics.
Remarkable progresses have been achieved while the existing methods typically treat it as a direct regression or iterative refinement problem and often struggling to capture complex geometric transformations or generalize across different domains. 
In this work, we propose \textbf{HomoFM}, a new framework that introduces the flow matching technique from generative modeling into the homography estimation task for the first time.
Unlike the existing methods, we formulate homography estimation problem as a velocity field learning problem.
By modeling a continuous and point-wise velocity field that transforms noisy distributions into registered coordinates, the proposed network recovers high-precision transformations through a conditional flow trajectory.
Furthermore, to address the challenge of domain shifts issue, e.g., the cases of multimodal matching or varying illumination scenarios, we integrate a gradient reversal layer (GRL) into the feature extraction backbone.
This domain adaptation strategy explicitly constrains the encoder to learn domain-invariant representations, significantly enhancing the network's robustness.
Extensive experiments demonstrate the effectiveness of the proposed method, showing that HomoFM outperforms state-of-the-art methods in both estimation accuracy and robustness on standard benchmarks.
Code and data resource are available at https://github.com/hmf21/HomoFM.
\end{abstract}

\section{Introduction}

Homography estimation aims to model the planar projective transformation between two images.
It plays an essential role for computer vision and robotics applications, such as image fusion~\cite{ying2021unaligned,xu2023murf}, image stitching~\cite{zhao2021image,nie2022deep}, visual odometry~\cite{wang2021tt,xu2025cuahn}, visual tracking~\cite{yao2021learning} and visual geo-localization~\cite{wang2023fine,he2024leveraging,xiao2025uasthn}.
Traditional methods for homography estimation typically follows either direct photometric-based or sparse feature-based paradigms~\cite{szeliski2007image}.
Direct photometric-based methods exhibit robustness in low-texture regions, however their may struggle in the case of large displacements.
On the other hand, feature-based approaches handle large motion effectively yet remain sensitive to the quality of correspondences especially in degraded or low-texture scenarios.

Early deep learning-based homography estimation method~\cite{detone2016deep} leverages convolutional neural networks (CNNs) to directly regress the corresponding geometric transformations from paired images and typically parameterize the homography through corner offsets.
Further efforts have been made on enhancing estimation accuracy by optimizing extracted features~\cite{nguyen2018unsupervised,le2020deep,zhang2020content,shao2021localtrans} or employing cascaded inference pipelines~\cite{cao2022iterative,zhu2024mcnet}.
However, these approaches often fail to maintain representation consistency during feature extraction, particularly when facing significant scale variations and domain shifts.
Furthermore, the reliance on recurrently stacked architectures inherently leads to the increase of parameter volume and computational overhead.
On the other hand, iterative architectures inspired by the inverse compositional Lucas-Kanade (IC-LK) algorithm~\cite{baker2004lucas} seek to progressively refine the predicted transformation through continuous updates of the homography matrix~\cite{chang2017clkn,zhao2021deep,cao2023recurrent}.
However, these methods suffer from slow convergence for the case of large displacement scenes.
Consequently, such a class of methods requires a substantial number of iterations to recover from suboptimal initializations, leading to excessive computational overhead and residual misalignment.
Recent work~\cite{zhang2025adapting} incorporates pretrained feature from foundation model DINOv2~\cite{oquab2023dinov2} with grid-based aligning strategy to achieve both the high accuracy and good efficiency.
However, these methods heavily rely on the discriminative semantic information provided by large foundation model, incurring significant computational costs.
Therefore, it suffers from huge parameter costs and struggles to generalize well when encountering significant domain shifts.

\begin{figure*}[t]
    \centering
    \includegraphics[width=\linewidth]{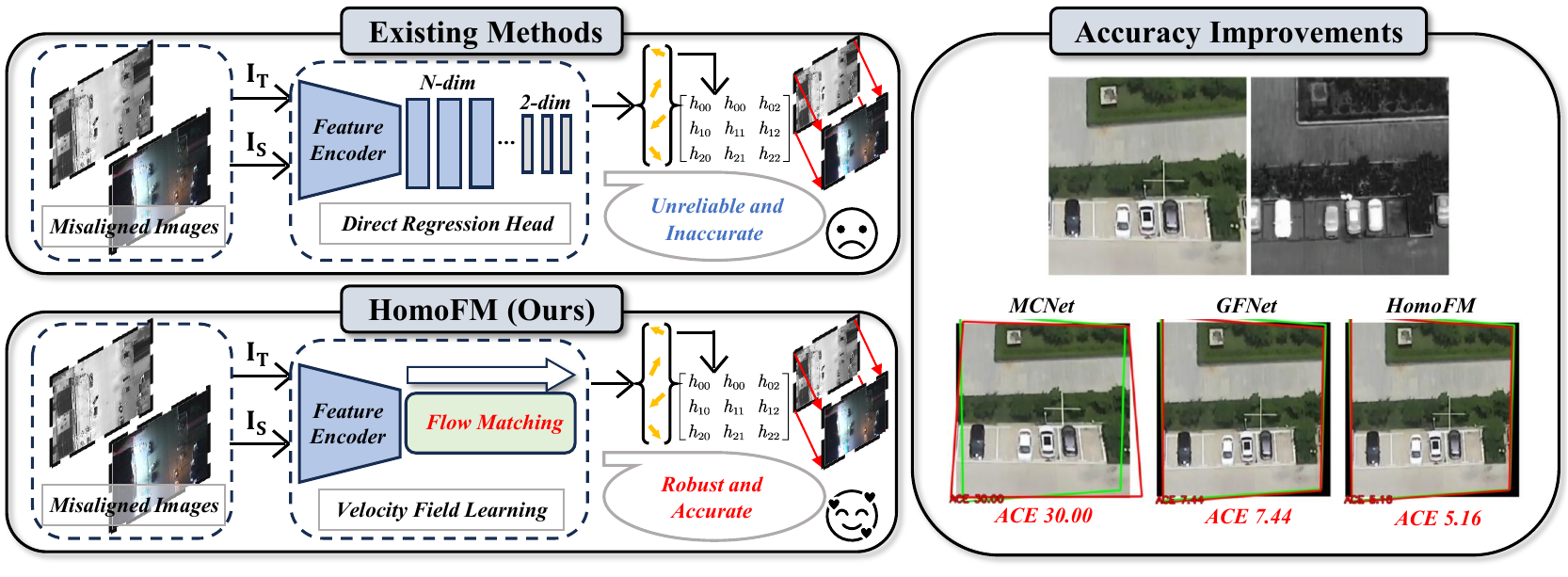}
    \vspace{-3.0ex}
    \caption{The differences between existing homography estimation methods and the proposed model.}
    \vspace{-3.0ex}
    \label{fig:verification_dataset}
\end{figure*}

Recent advances in generative modeling, particularly flow matching (FM) technique~\cite{lipman2022flow,liu2022flow}, have been successfully extended to widespread applications beyond pure image generation task~\cite{gui2025depthfm,sun2025rectified,pan2025register}.
Thanks to its capability to construct optimal transport paths with straighter trajectories, flow matching exhibits superior inference efficiency and precise response to conditional inputs, ensuring that the generated results faithfully align with the expected targets. 
Standard pipelines for homography estimation problem~\cite{cao2023recurrent,zhu2024mcnet,zhang2025adapting} directly regress static displacement vectors from extracted features using CNN or ViT blocks, where stacked layers rigidly compress high-dimensional feature representations into simple 2D displacement information.
Such a mapping strategy overlooks the continuous dynamics of geometric transformations, treating the alignment as a direct step rather than a controllable, progressive process.

In this paper, we propose \textbf{HomoFM}, a novel homography estimation method based on flow matching technique.
Unlike existing methods, HomoFM formulates a conditional velocity field that captures the intrinsic continuity of homography transformations.
This enables more expressive representation learning and ensures consistent performance under severe domain gaps.
The main motivation of HomoFM is to enhance the estimation accuracy by modeling the displacement field through predicted velocity vectors that drive the point-wise geometric transformations.
The proposed method departs from points translation regression method in two ways:
(i) \textbf{Continuous trajectory modeling}: we cast the alignment task as a conditional generation process that recovers the continuous warping trajectory, rather than a direct regress progression.
This allows the model to capture the dynamic evolution of the geometric transformation;
and (ii) \textbf{Robustness against long-range displacements}: rather than regressing static displacement values directly from feature maps, we operate on the velocity field of the pixel flow.
This makes the model inherently robust to severe perspective changes and ambiguous matching that often degrade conventional CNN-based methods.
To further boost the estimation performance, we also add a domain classifier connected to the feature encoder, which can ensure the extracted features having similar distributions over two domains by adding a gradient reversal layer (GRL).
By integrating GRL into the backbone, our model learns from multi-domain datasets, yielding transferable feature representations that generalize effectively across diverse scenarios, e.g., cross-modal matching tasks.
Experimental results on three datasets containing common and multimodal scenarios show HomoFM achieves an average improvement 2.31\% in AUC@3, compared to the previous SOTA GFNet~\cite{zhang2025adapting}.
Zero-shot evaluations also approve that HomoFM has stronger generalization than other homography estimation methods.

In summary, this paper has the following contributions:
\begin{itemize}
    \item We propose \textbf{HomoFM}, the first framework to formulate homography estimation as a generative modeling via flow matching. This paradigm achieves state-of-the-art alignment accuracy on multiple datasets, while reducing MACs by 91.10\% compared to the competing homography estimation method.
    \item We introduce a domain adversarial branch attached with gradient reversal layer (GRL) to enforce domain-invariant feature learning. Notably, this adversarial branch acts as a training-only constraint and is discarded during inference, providing a cost-free enhancement to cross-domain robustness without additional computational overhead.
    \item We validate the generalization of the proposed model on a newly collected zero-shot benchmark (AVIID-homo). Without any fine-tuning, HomoFM significantly outperforms existing methods in unseen scenarios, demonstrating its generalization to domain-invariant geometric constraints.
\end{itemize}

\section{Related Works}

\subsection{Deep Homography Estimation}

Deep homography estimation mainly follows two paradigms.
{(i) Regression-based estimation} predicts a compact transformation parameterization directly from image pairs.
A representative line regresses 4-point corner offsets and recovers the homography via DLT-style parameterization, e.g.,  HomographyNet~\cite{detone2016deep}.
To reduce supervision, unsupervised variants incorporate photometric reprojection and cycle-consistency objectives~\cite{nguyen2018unsupervised,zhang2020content}.
To further improve accuracy, cascaded and recurrent designs progressively refine the transformation through multi-stage inference~\cite{cao2022iterative,cao2023recurrent,zhu2024mcnet}.
At the same time, optimization-based formulations integrate Lucas-Kanade~\cite{baker2004lucas} style compositional updates into differentiable pipelines to ensure geometric consistency during refinement~\cite{baker2004lucas,chang2017clkn,lv2019taking,zhao2021deep}.
{(ii) Correspondence-driven estimation} strengthens homography estimation by predicting reliable matches and then solving the transformation.
Learned matcher and efficient filtering improve robustness under wide baselines and challenging appearance changes~\cite{sun2021loftr,lindenberger2023lightglue,yang2025multimodal}.
Dense matching methods further enhance robustness in low-texture regions by estimating near pixel-level correspondences (e.g., DKM~\cite{edstedt2023dkm} and RoMa)~\cite{edstedt2024roma}, although with higher computation.
Bridging dense correspondence and global regression, GFNet~\cite{zhang2025adapting} utilize DINOv2~\cite{oquab2023dinov2} representations to achieve efficient cross-domain alignment via structured displacement regression.
Despite those remarkable progresses, most of existing methods still cast alignment using direct displacement regression, while they may struggle with large displacement and ambiguous cross-modal scenarios.
This motivates us to consider to solve this problem from a flow-matching-based perspective that explicitly model alignment as a transport process via stage-wise updates.

\subsection{Flow Matching for Vision Tasks}
Flow Matching (FM) serves as an efficient alternative to diffusion models.
By regressing ODE-defined vector fields, it effectively transports samples along smooth probability paths.
In practical applications, these generated trajectories are often effective and more stable than stochastic diffusion paths, enabling fast inference with few function evaluations~\cite{lipman2022flow,liu2022flow}.
Compared to traditional diffusion models, FM is particularly attractive for vision tasks that require efficient sampling or structured generation.

Recent progress has extended FM beyond image synthesis to a broad range of vision tasks.
In 2D conditional prediction, FM has been applied to cross-modal generation such as image-to-depth estimation with few-step sampling~\cite{gui2025depthfm}.
In point cloud registration task, recent work formulates registration as conditional flow matching to better handle noise and large displacement~\cite{pan2025register}.
FM has also been adopted in vision-based decision making and planning, where flow-based policies and discrete flow matching algorithm for trajectory generation are proposed to improve sampling efficiency~\cite{gode2024flownav,chang2025efficientflow,xu2025wam}.
Despite these advances, applying FM to 2D geometric alignment, specifically homography estimation, remains underexplored.
In this work, we bridge this gap by introducing the first flow-matching-based framework for homography estimation task, leveraging FM's transport perspective to improve the performance for large-displacement and cross-modal challenges.

\section{Method}
\label{sec:method}
We propose HomoFM, a novel framework that reformulates homography estimation through the lens of conditional flow matching.
Rather than treating the alignment as a direct one-step regression task, we model the homography transformation as a deterministic, continuous transport process that progressively warps the source grid point to the target coordinate.

\subsection{Domain Adapted Feature Extraction}
\label{sec:feature_extraction}

\begin{figure}[b]
    \centering
    \includegraphics[width=\linewidth]{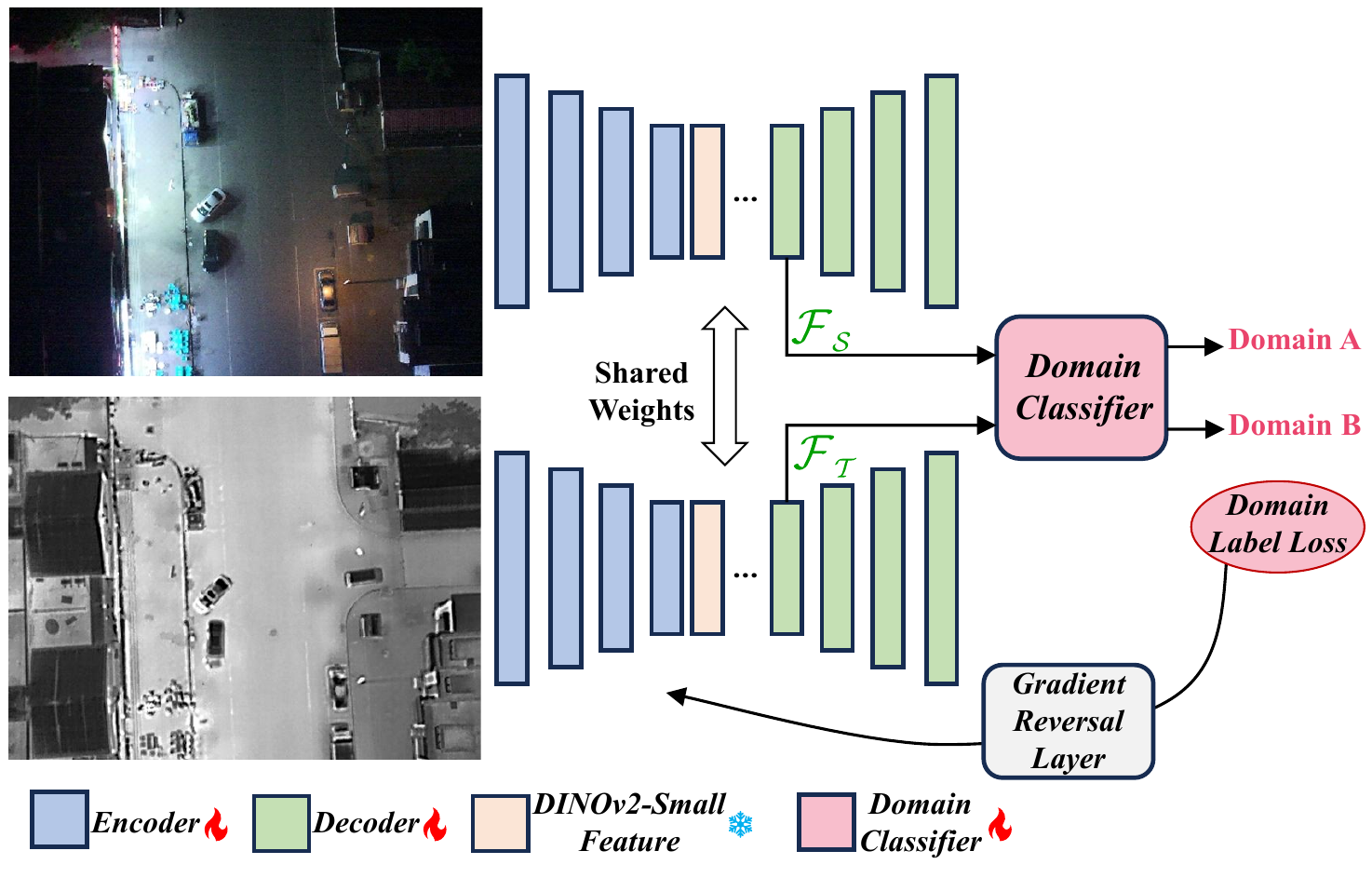}
    \vspace{-1.0ex}
    \caption{The overview of the domain adapted design in feature extraction, taking visible-infrared alignment problem as an example.}
    \label{fig:GRL}
\end{figure}

Given a source image $I_S$ and a target image $I_T$, our goal is to estimate an accurate homography transformation.
To extract discriminative features, we adopt the DINOv2-Small model~\cite{oquab2023dinov2} as our backbone.
This choice significantly improves computational efficiency compared to prior works like GFNet~\cite{zhang2025adapting} and RoMa~\cite{edstedt2024roma}, which rely on the heavier DINOv2-Large model.
Following the design of GFNet, we equip the backbone with a feature pyramid network (FPN) incorporating stacked cross-attention layers to reconstruct features across different spatial scales of $1/14$, $1/8$, $1/4$, $1/2$ and $1$.
To leverage strong semantic information, the extracted low-resolution DINOv2 features are up-sampled via interpolation to resolve the resolution mismatch and then concatenated with the lowest-level features of the FPN.
To ensure robustness against domain shifts, we explicitly integrate a gradient reversal layer (GRL)~\cite{ganin2015unsupervised} into our framework and the design of the whole feature extraction network is shown in Figure~\ref{fig:GRL}. 
During training, the GRL reverses the gradient signs flowing back to the backbone, enforcing the network to learn domain-invariant geometric features while suppressing domain-specific style information.

In particular, given the source image $I_S$ and target image $I_T$, which may originate from different distributions (e.g., visible vs. infrared), standard encoders tend to retain domain-specific signatures in the extracted features $\mathcal{F}_S$ and $\mathcal{F}_T$.
To avoid these modality-specific cues, we attach a domain discriminator $D$ to the feature decoder via the GRL as shown in Figure~\ref{fig:GRL}.
We assign domain labels $y_S=0$ to the source features and $y_T=1$ to the target features.
The domain discriminator $D$ aims to classify the origin of the features by minimizing the binary cross-entropy loss:
\begin{equation}
    \mathcal{L}_{dom} = - \left[ \log D(\mathcal{F}_S) + \log (1 - D(\mathcal{F}_T)) \right],
    \label{eq:grl_loss}
\end{equation}
where $D(\cdot)$ represents the discriminator's predicted probability of the input belonging to the source domain.
Crucially, during the inference, the GRL acts as an identity transform between the backbone and the discriminator and leaves the input unchanged.
During backpropagation, GRL reverses the gradient by a factor of $-\alpha$:
\begin{equation}
    \frac{\partial \mathcal{L}_{dom}}{\partial \mathcal{F}} \leftarrow -\alpha \frac{\partial \mathcal{L}_{dom}}{\partial \mathcal{F}}.
\end{equation}
where $\alpha$ is a given constant and $\mathcal{F}$ presents the extracted features ($\mathcal{F}_S$ or $\mathcal{F}_T$).
This reversal forces the backbone to maximize $\mathcal{L}_{dom}$, thereby learning domain-invariant representations where extracted features become indistinguishable to the discriminator, only containing the shared geometric structures.
To preserve the integrity of fine-grained geometric details, we specifically attach the domain discriminator to the first layer of the feature decoder.
This configuration restricts the domain adversarial training to the encoder and initial decoder. 
Consequently, early-stage features are guaranteed to be aligned across domains, while prevents detail loss during the subsequent dense matching process.

It is worth noting that this domain adversarial branch serves exclusively as a regularization constraint during the training process and can be completely discarded during inference.
This design allows the network to revert to its original efficient architecture for deployment.
Consequently, our method acts as a cost-free performance booster, significantly enhancing cross-domain robustness without introducing any additional computational overhead or memory footprint.

\subsection{Displacement Estimation with Flow Matching}
\label{sec:optimal_transport}

Homography estimation essentially aims to find a mapping that transports pixel coordinates from the source plane to the target plane.
The key problem is to solve the displacement field $\mathbf{w} \in \mathbb{R}^{H \times W \times 2}$, where the $H$ and $W$ are the height and width of the images.
With the estimated displacement field, the global homography matrix $\mathbf{H}$ can be subsequently derived via a least-squares fitting.
Previous works \cite{cao2023recurrent,zhu2024mcnet,zhang2025adapting} predict $\mathbf{w}$ in a multi-scale manner, using extracted features $\mathcal{F}_S$ and $\mathcal{F}_T$ from the coarsest level to the finest.
We argue that this alignment process is naturally equivalent to a time-dependent flow problem.
Specifically, the displacement of a point can be viewed as a particle moving along a trajectory from its initial position (identity grid) to its final position (aligned coordinate).
Based on this insight, we adopt the conditional flow matching (CFM)  framework~\cite{lipman2022flow} to model this geometric transport process.
Unlike generative tasks that start with Gaussian noise, homography estimation is a deterministic regression task with a unique ground truth displacement.
Consequently, we simplify the flow matching formulation by using a Dirac prior~\cite{li2025audio} centered at zero ($x_0 = \mathbf{0}$), which provides an informative initialization.
This formulation allows the model to fully exploit the prior displacement constraints of the task, bypassing the redundant denoising process.
This physically implies that at time $t=0$, the grid has zero displacement at the original coordinate.

We define the alignment trajectory as a transport path from the identity grid (zero displacement, $x_0 = \mathbf{0}$) to the target grid deformed by the ground truth displacement $x_1 = \mathbf{w}_{gt}$. 
Using optimal transport interpolation, the intermediate flow state $x_t$ at time $t,t \in [0, 1]$ is defined as:
\begin{equation}
    x_t = (1 - t) x_0 + t x_1 = t \cdot \mathbf{w}_{gt}; \quad\frac{d}{dt} x_t = \mathbf{w}_{gt}.
    \label{eq:interpolant}
\end{equation}
Specifically, $x_t$ denotes an intermediate displacement on the misaligned image plane, representing a partial transition towards the ground-truth displacement $\mathbf{w}_{gt}$. 
To recover this trajectory using the ODE-based flow matching framework, we formulate the optimization target by taking the time derivative of the current displacement.
This derivation implies that the rate of change is constant and equivalent to the total displacement.
Therefore, we parameterize this vector field using a network $v_\theta(x_t, t, C)$, which takes the current flow state $x_t$, time $t$, and the image context $C$ as inputs.
Notably, the image context $C$ consists of the concatenated features of $\mathcal{F}_S$ and $\mathcal{F}_T$ from the decoder presented in Section~\ref{sec:feature_extraction}.
To incorporate the time information, we utilize a FiLM-based residual head to embed time $t$, which effectively informs the network of the current progression stage (i.e., the scale of the input displacement).
During inference, we employ an Euler solver to recover displacement
\begin{equation}
    x_{t_{n}} = x_{t_{n-1}} + v_\theta(x_{t_{n-1}}, t_{n-1}, C) \cdot \Delta t,
    \label{eq:euler_step}
\end{equation}
where $\Delta t = 1/N$ is the time step size with $N$ being the given constant of the total steps, $t_n$ ($t_n=n\cdot\Delta t $) represents the current time embedding.
And $x_{t_n}$ presents the current state of displacement, which can be calculated from the state $x_{t_{n-1}}$ and velocity $v_\theta(x_{t_{n-1}}, t_{n-1}, C)$ of the previous step.
In addition, $\mathbf{x}_{0}$ is initialized as $\mathbf{0}$ following our Dirac prior.
The iterative process progressively guides the grid points to their optimal aligned coordinates, ensuring high precision for homography estimation.

The entire sequence of estimated displacements $\mathbf{w}_{N}=x_{t_N}$ generated by the ODE solver is designed as training objective.
We employ the $L_2$ loss between the predicted displacement $\mathbf{w}_{N}$ and the ground truth grid coordinate $\mathbf{w}_{gt}$ as the alignment loss $\mathcal{L}_{FM}$.
\begin{equation}
    \mathcal{L}_{FM} = \rho(\| \mathbf{w}_{N} - \mathbf{w}_{gt}\|_2),
    \label{eq:euler_step}
\end{equation}
where $\rho(\cdot)$ is a cost function from RoMa~\cite{edstedt2024roma}.
Combined with the domain classification loss from the aforementioned GRL module, the total objective is defined as $\mathcal{L} = \mathcal{L}_{FM} + \lambda \mathcal{L}_{dom}$.
Moreover, we also adopt the supervision strategy for multi-scale features given in GFNet~\cite{zhang2025adapting} to compute $\mathcal{L}_{FM}$.
This involves aggregating the estimation errors across different hierarchical levels of the FPN to ensure coarse-to-fine alignment accuracy.


\begin{table*}[t]
\centering
\caption{
We compare the proposed HomoFM with existing SOTA methods on three datasets. We report both alignment accuracy and efficiency metrics. (\textbf{bold}: best, \underline{underline}: second) (Group A: Image matching. Group B: Homography estimation.)
}
\label{tab:comparison}
\renewcommand{\arraystretch}{1}

{
\begin{tabular}{|c|c|cccc|cccc|}
\hline
\multirow{2}{*}{Group} & \multirow{2}{*}{Method} & \multicolumn{4}{c|}{MSCOCO} & \multicolumn{4}{c|}{VIS-IR} \\
 & & auc@3$\uparrow$ & auc@5$\uparrow$ & auc@10$\uparrow$ & auc@20 & auc@3$\uparrow$ & auc@5$\uparrow$ & auc@10$\uparrow$ & auc@20$\uparrow$ \\
\hline
\multirow{4}{*}{A} 
 & SIFT+LightGlue & 89.43 & 93.48 & 96.64 & 98.27 & 2.56 & 8.91 & 23.33 & 39.99 \\
 & LoFTR & 94.19 & 96.51 & 98.25 & 99.12 & 10.17 & 22.51 & 45.93 & 65.84 \\
 & ELoFTR & 95.65 & 97.39 & 98.69 & 99.34 & 13.28 & 26.32 & 49.43 & 69.10 \\
 & RoMa & 96.01 & 97.61 & 98.80 & 99.40 & \textbf{25.74} & \textbf{39.79} & \textbf{62.49} & \textbf{80.14} \\
\hline
\multirow{5}{*}{B} 
 & RHWF & 93.02 & 95.81 & 97.90 & 98.77 & 18.16 & 32.05 & 54.48 & 72.84 \\
 & MCNet & 91.23 & 94.73 & 97.36 & 98.68 & 16.10 & 30.59 & 51.51 & 71.82 \\
 & PRISE & 88.64 & 92.83 & 96.23 & 97.14 & 13.28 & 24.20 & 48.60 & 66.60 \\
 & GFNet & \underline{97.69} & \underline{98.61} & \underline{99.30} & \underline{99.65} & {21.28} & {35.72} & {58.44} & \underline{76.23} \\
& {\cellcolor{gray!15}\textbf{HomoFM (ours)}} & 
\cellcolor{gray!15}\textbf{98.45} & \cellcolor{gray!15}\textbf{99.07} & \cellcolor{gray!15}\textbf{99.54} & \cellcolor{gray!15}\textbf{99.77} & 
\cellcolor{gray!15}\underline{24.51} & \cellcolor{gray!15}\underline{37.94} & \cellcolor{gray!15}\underline{58.82} & \cellcolor{gray!15}{75.95} \\
\hline
\end{tabular}
}

\vspace{6pt}

{%
\begin{tabular}{|c|c|cccc|cc|cc|}
\hline
\multirow{2}{*}{Group} & \multirow{2}{*}{Method} & \multicolumn{4}{c|}{GoogleMap} & & & \multicolumn{2}{c|}{Efficiency} \\
 & & auc@3$\uparrow$ & auc@5$\uparrow$ & auc@10$\uparrow$ & auc@20$\uparrow$ & & & Params (MB)$\downarrow$ & MACs (G)$\downarrow$ \\
\hline
\multirow{4}{*}{A} 
 & SIFT+LightGlue & - & - & - & - & & & 11.88 & 38.51 \\
 & LoFTR & 4.99 & 13.87 & 32.27 & 51.24 & & & 11.56 & 249.7 \\
 & ELoFTR & 5.37 & 14.49 & 33.60 & 52.73 & & & 16.02 & 179.8  \\
 & RoMa & 45.29 & 62.18 & 78.63 & 88.23 & & & 111.28 & 2503  \\
\hline
\multirow{5}{*}{B} 
 & RHWF & 18.32 & 37.47 & 61.09 & 76.43 & & & \underline{1.29} & \underline{23.43} \\
 & MCNet & 16.31 & 35.28 & 59.51 & 75.63 & & & \textbf{0.85} & \textbf{4.56} \\
 & PRISE & 11.41 & 31.68 & 54.78 & 70.54 & & & 19.24 & 55.25 \\
 & GFNet & \underline{51.44} & \underline{66.34} & \underline{80.62} & \underline{89.66} & 
 & & 3.86 & 1657 \\
& {\cellcolor{gray!15}\textbf{HomoFM (ours)}} & 
\cellcolor{gray!15}\textbf{54.39} & \cellcolor{gray!15}\textbf{68.65} & \cellcolor{gray!15}\textbf{82.01} & \cellcolor{gray!15}\textbf{89.74} &
& & \cellcolor{gray!15}{3.37} & \cellcolor{gray!15}{147.4} \\
\hline
\end{tabular}
}
\end{table*}

\section{Experiments}

\subsection{Datasets and Experimental Setup}
We evaluate the proposed HomoFM on three widely used datasets covering both natural and multimodal scenarios.
First, for standard homography estimation, we select the MSCOCO dataset~\cite{lin2014microsoft}, which features diverse scenes and serves as a fundamental benchmark on this field~\cite{zhao2021deep,cao2022iterative,zhu2024mcnet}. 
To ensure consistency, we follow the experimental setting of GFNet~\cite{zhang2025adapting} by training on the same generated basic datasets and evaluating on the standard MSCOCO dataset.
Second, following the same setting as previous works~\cite{cao2022iterative,cao2023recurrent,zhao2021deep,zhang2025adapting}, we employ the satellite-map and visible-infrared datasets to verify the cross-domain capabilities of HomoFM.
We adopt the same training and testing splits provided by~\cite{zhang2025adapting} to ensure a fair and direct comparison with state-of-the-art homography estimation methods.

Crucially, given the iterative nature of the flow matching solver, we apply gradient clipping with a maximum norm of $1.0$ to prevent gradient explosion and ensure numerical stability during mixed-precision training.
We set the number of flow matching steps to $N=4$, to balance inference efficiency and estimation accuracy.
The weight $\lambda$ for the domain adversarial loss is set to be $0.01$.
To ensure the stable convergence of the primary alignment task, we employ a warm-up strategy where the GRL branch is deactivated during the initial $5\%$ of the training iterations.
All training and evaluation experiments are conducted on a single NVIDIA A100 GPU (40GB).
We assess the alignment accuracy using the standard average corner error (ACE) and the area under the curve (AUC) of ACE at error thresholds of 3, 5, 10 and 20 pixels.
Furthermore, model efficiency is measured in terms of the number of learnable parameters and multiply-accumulate operations (MACs).

\subsection{Evaluation Results}
To demonstrate the effectiveness of our approach, we conduct a comprehensive comparison with state-of-the-art methods and report the results in Table~\ref{tab:comparison}.
The comparative baselines are categorized into two groups: feature matching baselines (Group A) (including SIFT~\cite{lowe2004distinctive}+LightGlue~\cite{lindenberger2023lightglue}, LoFTR~\cite{sun2021loftr}, ELoFTR~\cite{wang2024efficient} and RoMa~\cite{edstedt2024roma}) and deep homography methods (Group B) (including RHWF~\cite{cao2023recurrent}, MCNet~\cite{zhu2024mcnet}, PRISE~\cite{zhang2023prise} and GFNet~\cite{zhang2025adapting}).
For a fair and consistent comparison, we adopt the standardized evaluation protocol established in~\cite{zhang2025adapting}.
Results for several baseline methods are directly obtained from their reported benchmarks, where all methods are conducted under identical training and testing configurations.
As shown in Table~\ref{tab:comparison}, HomoFM demonstrates superior accuracy across diverse scenarios, consistently outperforming existing deep homography methods while achieving highly competitive results against feature matching baselines.

\begin{figure}[t]
    \centering
    \includegraphics[width=\linewidth]{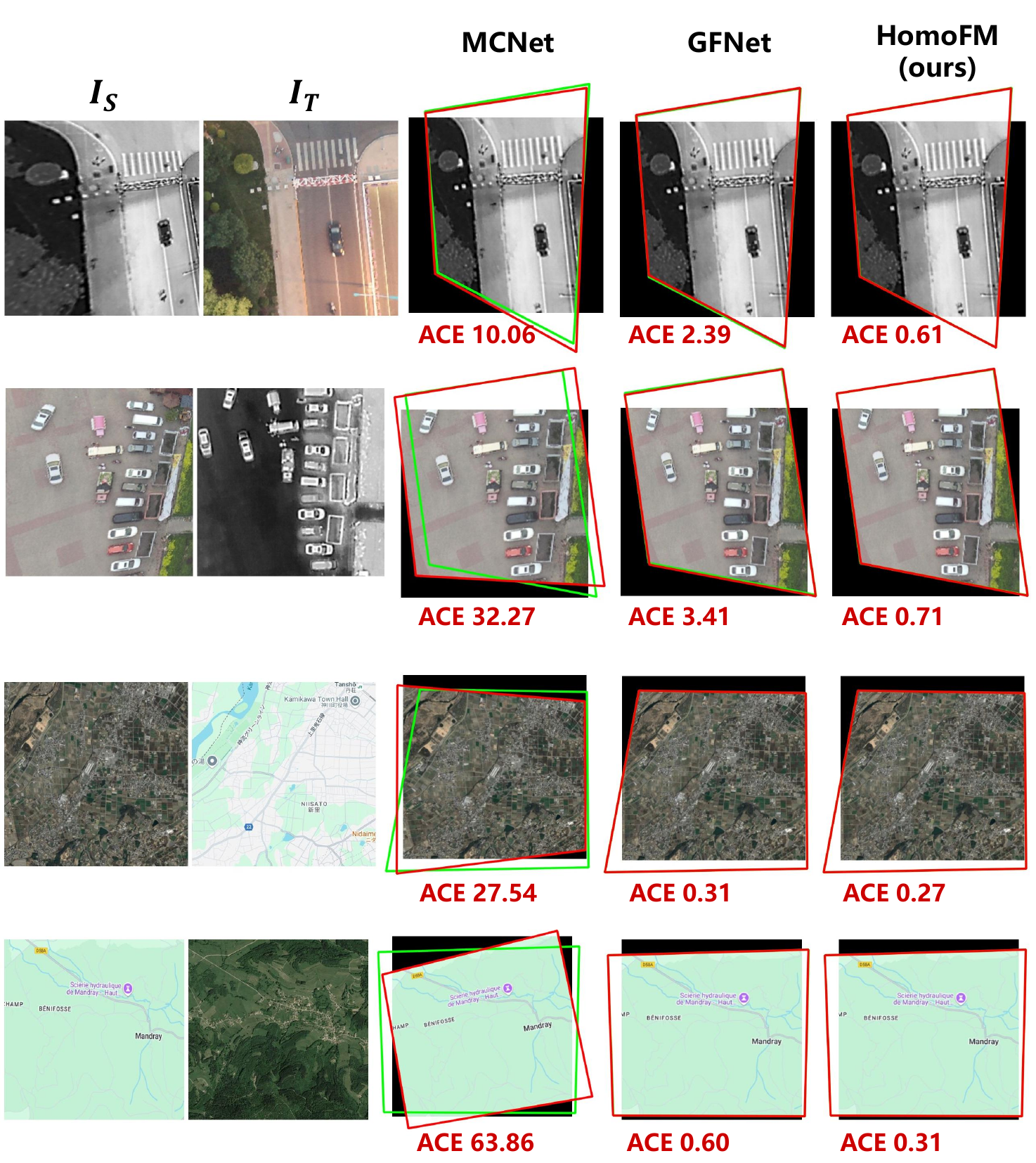}
    \vspace{-5.0ex}
    \caption{Qualitative comparison of homography estimation results. The green and red polygons represent the ground truth and predicted alignment of $I_S$ onto $I_T$, respectively. The ACE is annotated on each sample for quantitative reference.}
    \label{fig:match_viz}
    \vspace{-2.0ex}
\end{figure}

In particular, on MSCOCO benchmark, HomoFM sets a new state-of-the-art with an AUC@3 of 98.45\%, surpassing the previous best method GFNet (97.69\%) and the dense matcher RoMa (96.01\%).
This outperforming score indicates that the proposed method successfully recovers pixel-level alignment details.
In the VIS-IR (visible-infrared) dataset, HomoFM continues to lead within the homography estimation category (Group B), surpassing GFNet by \textbf{+3.23\%} in AUC@3.
Also, although RoMa achieves higher AUC@3 of 25.74\%, HomoFM presents a distinct advantage in terms of parameter count and memory footprint, as further detailed in the subsequent efficiency analysis.
Aside from the heavy-weight RoMa model, our method distinctly outperforms other sparse and semi-dense matching models.
This highlights the superior capability of our method in adapting to low-light and infrared cross-modal data.
For the GoogleMap (satellite-map) dataset, feature-matching methods like RoMa struggle as there are significant appearance differences between satellite and map views.
In contrast, HomoFM achieves 54.39\% AUC@3, outperforming RoMa by a significant margin of +9.10\%.
Compared to GFNet model, we also achieve a solid improvement of +2.95\% in AUC@3 and +2.31\% in AUC@5.
In summary, HomoFM achieves the significant gains particularly at AUC@3 and AUC@5, highlighting its advantage in delivering high-precision alignment results compared to previous methods.
Some comparison samples are also presented in Figure~\ref{fig:match_viz}.

Beyond superior accuracy, HomoFM demonstrates remarkable efficiency and stands out as an ideal deployment solution for time-critical scenarios.
In particular, as shown in the efficiency evaluation separately listed in Table~\ref{tab:comparison}, our method maintains a lightweight design with only 3.37M learnable parameters.
This is slightly fewer than GFNet and significantly smaller than the large-scale dense matcher RoMa.
More importantly, benefiting from the efficient DINOv2-Small backbone, HomoFM largely reduces the computational overhead.
Our model requires only 147.4G MACs, representing a substantial 91\% reduction compared to GFNet (1657G) and a 94\% reduction compared to RoMa (2503G).
In summary, although HomoFM does not rank first in efficiency metrics (e.g., compared to lightweight regression baselines), it still achieves a wonderful trade-off between high-precision alignment and low computational resource consumption.

\subsection{Ablation Studies}

\subsubsection{Flow Matching Method}
While above results have shown the overall advantages of our framework, we conduct additional controlled experiments to isolate the impact of the flow matching method.
To ensure a fair comparison, we perform an ablation study focusing solely on the regression head with all other components fixed.
In particular, we fix the feature encoder as the DINOv2-Small model and train all these models without the aforementioned GRL module.
For the direct regression pipeline, we construct convolutional heads with different numbers of hidden blocks to investigate the impact of model capacity. 
We select two representative datasets for the evaluation: MSCOCO for standard scenes, and GoogleMap for challenging multimodal tasks.
The evaluation results are summarized in Table~\ref{tab:ablation_FM}.
It can be observed that simply increasing the depth of the convolutional regression head from 8 to 16 hidden blocks yields only marginal improvements.
Further scaling the regression head fails to deliver proportional accuracy gains, while leading to unnecessary computational overhead.
In contrast, replacing the regression head with the proposed flow matching strategy leads to a substantial performance boost.
This confirms that the gains in alignment accuracy are derived from the flow matching methodology, which can hardly be obtained by simply increasing model parameters.
We also investigate the impact of the ODE solver's step count $N$ on alignment performance.
By comparing the single-step inference ($N=1$) against the default setting of $N=4$ on the GoogleMap dataset, we observe that the AUC@3 drops from 53.41\% to 51.21\% and the AUC@5 drops from 68.16\% to 66.11\%.
This result also highlights the accuracy gains attributed to the multi-step refinement and a step-reduced HomoFM still maintains competitive performance compared to the direct regression baselines.

\begin{table}[b]
\centering
\caption{Ablation study on flow matching module. We report \textbf{auc@3} ($\uparrow$) and \textbf{MACE} ($\downarrow$) across three datasets. The hidden 8 and 16 indicate the convolutional regression head has 8 and 16 hidden blocks.}
\label{tab:ablation_FM}
\renewcommand{\arraystretch}{1.3}
\resizebox{\linewidth}{!}
{
\begin{tabular}{|c|cc|cc|}
\hline
\multirow{2}{*}{Method} & \multicolumn{2}{c|}{MSCOCO} & \multicolumn{2}{c|}{GoogleMap} \\
\cline{2-5}
 & auc@3 $\uparrow$ & MACE $\downarrow$ & auc@3 $\uparrow$ & MACE $\downarrow$ \\
\hline
DINO-Small + Conv(hidden-8) & 97.33 & 0.08 & 48.34 & 3.03 \\
DINO-Small + Conv(hidden-16) & 97.66 & 0.07 & 48.55 & 2.98 \\
\rowcolor{gray!15} 
\textbf{DINO-Small + FM} & \textbf{98.35} & \textbf{0.05} & \textbf{53.41} & \textbf{2.83} \\
\hline
\end{tabular}
}
\end{table}

\subsubsection{Domain Adapted Feature}
We further investigate the effectiveness of the proposed domain-invariant feature extraction method.
To demonstrate the effectiveness of this module, we evaluate the impact of the domain adversarial under two distinct configurations: the standard convolutional regression baseline and the proposed flow matching framework. 
This setup allows us to verify whether the domain adaptation strategy yields consistent improvements independent of the downstream matching head.
For simplified comparison, experiments are conducted on MSCOCO and GoogleMap dataset, with quantitative results detailed in Table~\ref{tab:ablation_GRL}.
It is evident that integrating the GRL module consistently improves performance across both settings. 
These results demonstrate that the domain-invariant features learned via the adversarial branch are beneficial regardless of whether the homography is estimated through standard regression or the proposed flow matching framework.
Consequently, the GRL operates as model-agnostic component, capable of enhancing feature robustness in multimodal alignment tasks without relying on a specific network architecture.

\begin{table}[t]
\centering
\caption{Ablation study on the domain adapted feature design.}
\label{tab:ablation_GRL}
\resizebox{\linewidth}{!}
{
\renewcommand{\arraystretch}{1.3}
\begin{tabular}{|c|c|ccccc|}
\hline
\multirow{2}{*}{FM} & \multirow{2}{*}{GRL} & \multicolumn{5}{c|}{MSCOCO} \\
 & & auc@3$\uparrow$ & auc@5$\uparrow$ & auc@10$\uparrow$ & auc@20$\uparrow$ & MACE$\downarrow$ \\
\hline
\multirow{2}{*}{w/o} & w/o & 97.33 & 98.40 & 99.20 & 99.60 & 0.081 \\
& \cellcolor{gray!15}{with} & \textbf{97.50} & \textbf{98.50} & \textbf{99.25} & \textbf{99.63} & \textbf{0.075}\\
\hline
\multirow{2}{*}{with} & w/o & 98.35 & 99.01 & 99.51 & 99.75 & 0.050 \\ 
& \cellcolor{gray!15}{with}  &  \textbf{98.45} & \textbf{99.07} & \textbf{99.54} & \textbf{99.77} & \textbf{0.047}\\
\hline
\end{tabular}
}
\vspace{4pt}
\resizebox{\linewidth}{!}
{
\renewcommand{\arraystretch}{1.3}
\begin{tabular}{|c|c|ccccc|}
\hline
\multirow{2}{*}{FM} & \multirow{2}{*}{GRL} & \multicolumn{5}{c|}{GoogleMap} \\
 & & auc@3$\uparrow$ & auc@5$\uparrow$ & auc@10$\uparrow$ & auc@20$\uparrow$ & MACE$\downarrow$ \\
\hline
\multirow{2}{*}{w/o} & w/o & 48.34 & 64.20 & 78.96 & 87.91 & 3.03 \\
& \cellcolor{gray!15}{with} & \textbf{48.99} & \textbf{64.52} & \textbf{79.04} & \textbf{87.92} & \textbf{2.99}\\
\hline
\multirow{2}{*}{with} & w/o & 53.41 & 68.16 & 81.54 & 89.33 & 2.83 \\ 
& \cellcolor{gray!15}{with}  &  \textbf{54.39} & \textbf{68.65} & \textbf{82.01} & \textbf{89.74} & \textbf{2.77}\\
\hline
\end{tabular}
}
\end{table}

\subsection{Zero-shot Generalization Analysis}
\begin{figure}[t]
    \centering
    \includegraphics[width=\linewidth]{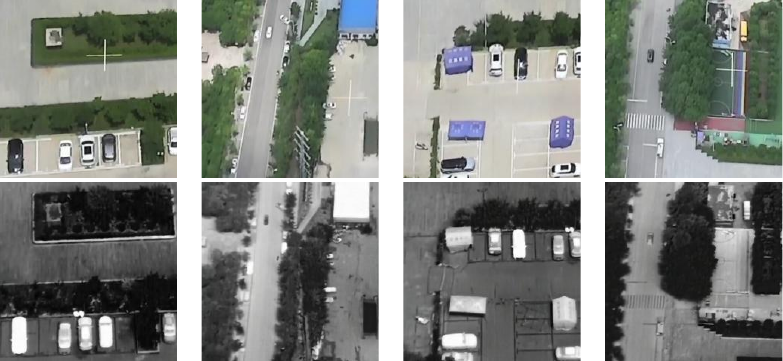}
    \vspace{-1.0ex}
    \caption{Representative test samples from the challenging AVIID-homo dataset. These pairs feature significant cross-modal discrepancies and are used to evaluate the model's zero-shot generalization.}
    \label{fig:zero_shot_samples}
\end{figure}

As discussed in previous sections, most existing deep homography estimation works are subject to the case where the training and testing sets are split from the same dataset.
While such a way demonstrates the model's capability to fit a specific data distribution, it fails to adequately assess its generalization capability to the unseen scenarios.
In real-world applications, however, models are often deployed in environments that differ significantly from their training data.
To further verify the robustness of the proposed model in a zero-shot setting, we construct a new test set derived from the AVIID dataset~\cite{han2023aerial}.
Specifically, we collect 1,000 visible-infrared pairs from AVIID and generate synthetic homography samples following the data generation method described in~\cite{zhang2025adapting}.
We maintain the same resolution setting ($448\times448$) and term the new dataset as AVIID-homo shown in Figure~\ref{fig:zero_shot_samples}.
Crucially, we directly evaluate the proposed HomoFM and other competing homography estimation methods (MCNet and GFNet) on these samples using weight pretrained on the VIS-IR dataset and without any fine-tuning.

The quantitative results are summarized in Table~\ref{tab:zero_shot}.
In contrast to the performance on the standard VIS-IR dataset (Table~\ref{tab:comparison}), all methods exhibit a substantial decrease in estimation accuracy.
For instance, the AUC@3 for all methods drops to near zero (less than 1\%).
This sharp performance drop highlights the severe challenge of zero-shot generalization, proving that pretrained models struggle in real-world applications.
It also emphasizes the necessity of evaluating generalization ability, which is often ignored in the existing benchmark.
Despite the challenging evaluation conditions, HomoFM still maintains a distinct lead over competing methods.
We achieve an AUC@3 of 0.70\% and AUC@5 of 13.58\%, significantly outperforming MCNet and GFNet.
Furthermore, our method reduces the MACE error to 4.71, which is also remarkably lower than other baselines.
All those results together demonstrate that the proposed flow matching module, combined with the domain-adversarial learning strategy, enables the network to capture intrinsic geometric representations.

\begin{table}[t]
\centering
\caption{Zero-shot evaluations for different homography estimation methods.}
\label{tab:zero_shot}
\renewcommand{\arraystretch}{1.3}
\resizebox{\linewidth}{!}
{
\begin{tabular}{|c|ccccc|}
\hline
\multirow{2}{*}{Method} & \multicolumn{5}{c|}{AVIID-homo} \\
 & auc@3$\uparrow$ & auc@5$\uparrow$ & auc@10$\uparrow$ & auc@20$\uparrow$ & MACE$\downarrow$ \\
\hline
MCNet & 0.12 & 1.74 & 10.91 & 22.87 & 30.89 \\
\hline
GFNet & 0.42 & 11.33 & 49.78 & 74.63 & 5.08\\
\hline
\rowcolor{gray!15} 
HomoFM & \textbf{0.70} & \textbf{13.58} & \textbf{53.22} & \textbf{76.50} & \textbf{4.71} \\
\hline
\end{tabular}
}
\end{table}

\section{Conclusion}
In this paper, we present HomoFM, the first framework to reformulate homography estimation problem as a deterministic transport problem via flow matching.
By modeling the alignment process as a linear trajectory starting from a zero-initialized grid, our method effectively transforms the complex generation task into a stable, iterative displacement regression problem. 
To further enhance robustness against domain shifts, we integrate a training-only GRL strategy, which enforces the learning of domain-invariant features without introducing any inference overhead.
Extensive experiments demonstrate that HomoFM achieves a new state-of-the-art on both standard and multimodal benchmarks, delivering high-precision alignment while reducing computational costs (MACs) by over 90\% compared to dense matching baselines. 
Moreover, its superior performance on the newly collected zero-shot benchmark highlights its exceptional generalization capability in unseen scenarios.







\newpage
\bibliographystyle{named}
\bibliography{ijcai26}

\end{document}